\documentclass[10pt,twocolumn,letterpaper]{article}

\usepackage{cvpr}              

\usepackage{graphicx}
\usepackage{amsmath}
\usepackage{amssymb}
\usepackage{booktabs}
\usepackage{multirow}
\usepackage{bm}
\usepackage{makecell}

\usepackage[pagebackref,breaklinks,colorlinks]{hyperref}

\usepackage[capitalize]{cleveref}
\crefname{section}{Sec.}{Secs.}
\Crefname{section}{Section}{Sections}
\Crefname{table}{Table}{Tables}
\crefname{table}{Tab.}{Tabs.}

\usepackage[marginal]{footmisc}

\def\confName{CVPR}
\def\confYear{2023}

\begin{document}
\title{InterPruner: Interactive Structured Pruning via Taylor-Implicit Criterion and
Language-Prior Modulator for Multimodal Object Detection}


\author{
  Qi Ming$^{1}$, 
  Zihan Yang$^{2}$, 
  Shaoguang Huang$^{3}$, 
  Si Sun$^{4}$, 
  Hanqing Zhang$^{5}$, \\
  Nanqing Liu$^{6}$, 
  Jiahui Lv$^{7}$, 
  Juan Fang$^{1}$, 
  Aleksandra Pizurica$^{8}$
  \\
  \small
  $^{1}$Beijing University of Technology, \small
  $^{2}$Southwest University, \small
  $^{3}$China University of Geosciences Wuhan \\\small
  $^{4}$Tsinghua University, \small
  $^{5}$Beijing Institute of Technology,
  $^{6}$Yunnan Normal University, \small
  $^{7}$Beijing Forestry University \\\small
  $^{8}$Department of Telecommunications and Information Processing (TELIN), Ghent University\small
  \\
  {\tt\small
  chaser.ming@gmail.com,~
  yangzihan0211@163.com,~
  Aleksandra.Pizurica@ugent.be
  }
}

\maketitle
\begin{abstract}
   Multimodal object detection proves effective in remote sensing, especially the RGB-Infrared paradigm.
The parallel feature extractors provide rich multimodal information for robust detection, yet introduce substantial channel redundancy and computational overhead. 
Existing pruning methods can reduce channel redundancy, but they are designed for unimodal backbones, overlooking cross-modal interactions and dynamic scene-wise redundancy. 
In this paper, we propose \textit{InterPruner}, the first interactive structured channel pruning framework for RGB–infrared object detectors.
Specifically, we first derive a Taylor-Implicit Criterion(TIC) to quantify channel importance via high-order Taylor expansion and implicit function theorem.
Then, a Modality Interaction Redundancy Analyzer (MIRA) identifies redundant channels via mutual compensability assessment.
Finally, a Scene-Prior Channel Anchor (SPCA) uses language priors as semantic anchors to measure channel-scene relevance for dynamic channel importance estimation.
Cross-modality channel pruning for RGB-Infrared detection is yet unexplored.
Extensive experiments on RGB-infrared object detection dataset demonstrate that InterPruner maintains high performance with negligible degradation. Specifically, it even achieves a 0.6\% mAP increase on the FLIR dataset when pruning 50\% of the channels. Code will be available on GitHub to facilitate future work.
\end{abstract}

\section{Introduction}
\label{sec:intro}

\begin{figure}[t]
    \centering
    \includegraphics[width=1\linewidth]{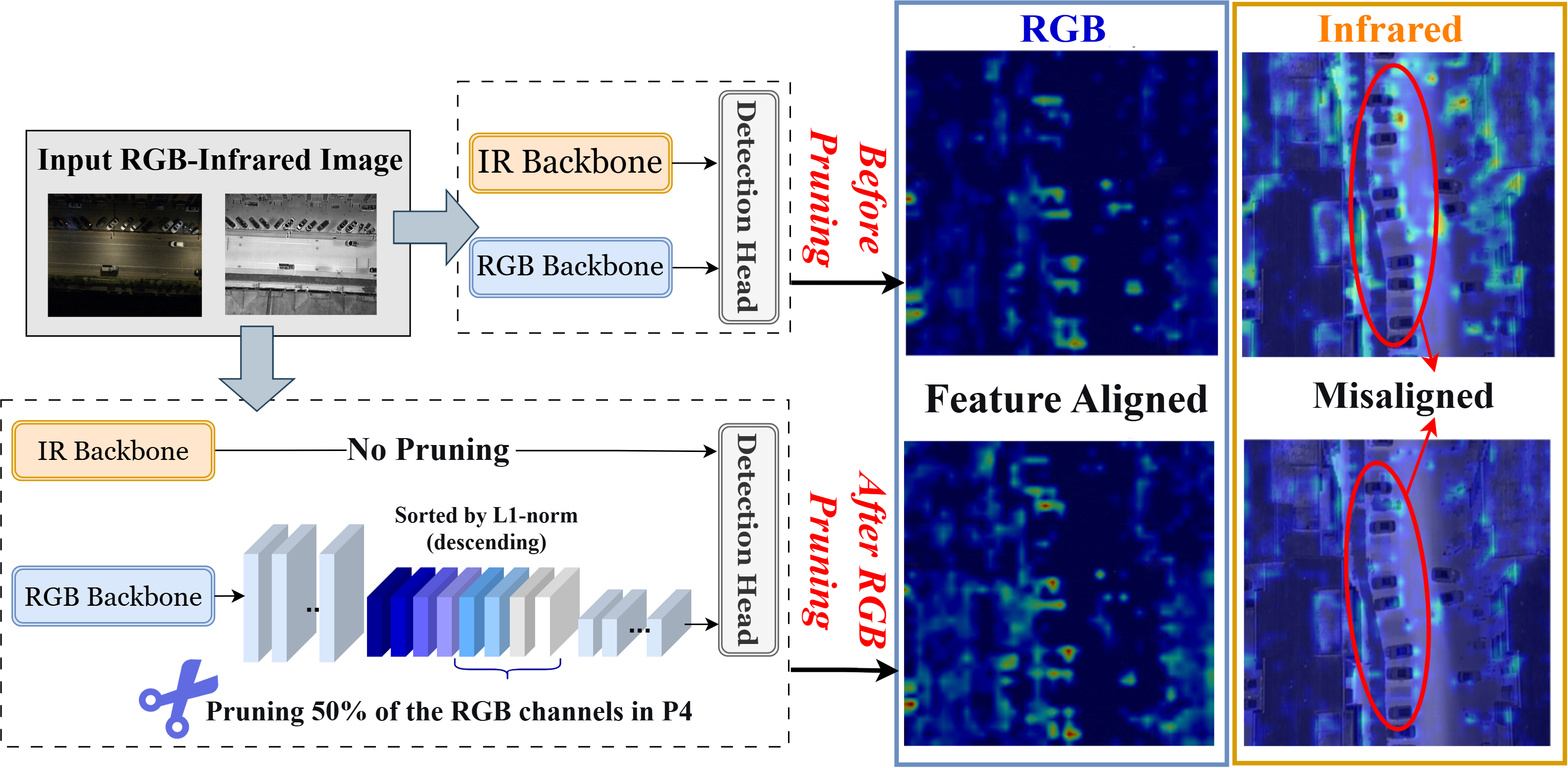}
    \caption{Feature visualizations of the RGB and Infrared branches before and after unimodal pruning at the P4 stage. Compared to the unpruned baseline (top), pruning the RGB branch (bottom) has a negligible impact on the RGB feature but leads to feature misalignment in the infrared branch.}
    \label{fig:inght}
\end{figure}

Multimodal object detection plays an important role in remote sensing applications\cite{SM3Det:li2025sm3detunifiedmodelmultimodal}.
By integrating visual cues from RGB\cite{zhu2025wavemamba}, infrared\cite{Thermal}, depth\cite{DepthWu_2025} and radar\cite{Radar}, multimodal systems seek to enhance scene understanding under adverse conditions. Among these modalities, RGB-infrared object detection\cite{shen_irdfusion_2026,yang2026crossweaver} serves as a representative application scenario. Specifically, RGB images provide rich texture and semantic information, while infrared images offer reliable temperature cues that are insensitive to illumination changes. Existing RGB-infrared detectors employ two modality-specific branches with feature fusion modules to extract features\cite{C2DFF-Net,AFFNet,wu2026bridgingrgbirgapconsensus}. 
However, the dual-branch architecture introduces substantial cross-modal redundancy, since many channels in the two streams contain correlated features that contribute little to detection accuracy. This redundancy causes unnecessary computation, making deployment on resource-constrained platforms challenging.

Pruning offers a promising solution to mitigate this efficiency problem via unstructured\cite{NEURIPS2025_b2c39fe6} or structured\cite{NetworkSliming} approaches. Structured pruning, in particular, operates at different granularities depending on the backbone: token-level\cite{VITPruning} for transformers and Mamba\cite{SSMPruning}, versus channel-level for convolutional neural networks\cite{BilevelPruning}.
Given the high resolution of remote sensing imagery, most RGB-Infrared detectors still adopt CNN backbones for their efficient multi-scale feature extraction. But these channel-level pruning techniques are mainly designed for single-stream CNNs. 

Directly extending single-stream pruning methods to the dual-stream setting results in performance degradation.
We attribute this degradation to two factors: 
\textbf{(1) Channel Importance is Coupled Across Modalities.} The two branches are inherently coupled via cross-modal interactions. Pruning one branch can impair the other's representation, even if the pruned channels appear unimportant in isolation.
\textbf{(2) Channel Importance Varies with Scenes.} Pruning decisions optimal for one scene may become suboptimal under another, as the redundancy distribution changes with environmental conditions\cite{wang2025efficienttesttimeadaptiveobject}. A fixed pruning criterion cannot adapt to such changes.

To overcome these limitations, we propose \textit{InterPruner}, the first interactive structured channel pruning framework for RGB–infrared object detection.
First, we formulate a Taylor-Implicit Criterion (TIC) that approximates the post-removal loss change through Taylor expansion and implicit function theorem to reveal each channel's genuine contribution.
Next, we employ a Modality Interaction Redundancy
Analyzer (MIRA) that alternately masks each modality and checks for compensation by the other to identify redundant channels.
Finally, a Scene-Prior Channel Anchor (SPCA) harnesses language priors to dynamically evaluate channel importance.
These modules produce a unified importance score that guides structured channel removal, preserving critical cross-modal information while adjusting to different scenes. InterPruner bridges the gap of structured pruning for RGB-infrared object detection, which remains unexplored to the best of our knowledge. 

In summary, the major contributions of our work are:
\begin{itemize}
\item We propose a Taylor-Implicit Criterion (TIC) that estimates channel importance via high-order loss expansion, providing a robust pruning basis for modalities.
\item A Modality Interaction Redundancy Analyzer (MIRA) is designed to identify cross-modal redundant channels via compensability assessment, ensuring that only informative features are retained for detection.
\item To make pruning decisions scene-aware, we introduce a Scene-Prior Channel Anchor (SPCA) that leverages scene priors, thereby enabling the model to identify redundancy across different visual contexts.
\end{itemize}

\section{Related Work}
\subsection{Multimodal Object Detection}
Multimodal detection integrates features from diverse sensors to enable robust perception. 
RGB-infrared detection, in particular, combines the visual cues with the thermal features, proving effective under adverse conditions\cite{AFFNet,C2DFF-Net}.
Various backbones have been explored for this task. 
Vision Transformers\cite{shen_irdfusion_2026} and their adapter-based variants\cite{Unirgb-ir} suffer from high computational cost. 
Emerging mamba-based models\cite{como:crossmamba} excel at global context modeling, yet their efficiency bottlenecks lie at the sequence level. In practice, CNNs\cite{AFFNet,C2DFF-Net,zhao2024removalselectioncoarsetofinefusion} remain the dominant backbone for RGB-Infrared detection, offering a favorable balance of efficiency and representational power. Despite their effectiveness, these dual-branch architectures introduce substantial channel redundancy, as many channels across the two streams encode correlated features with limited contribution to final detection accuracy. In this paper, we aim to reduce this redundancy and computational overhead while maintaining detection performance.

\subsection{Structured Channel Pruning}
Channel pruning removes entire filters or feature maps to achieve structured sparsity and produce compact models. 
Early approaches prune channels with small BN scaling factors\cite{NetworkSliming}, and later variants further separate important from unimportant channels via regularization \cite{li2022revisitingrandomchannelpruning}. More recent works attempt to estimate the post-pruning loss change via influence functions \cite{TaylorSecondorder} or unify static and dynamic pruning through bi-level optimization \cite{BilevelPruning}. 
Furthermore, domain-specific channel pruning has also been explored. IRPruneDeXt \cite{IRPruneDeXt} introduces wavelet-regularized channel pruning for infrared small target detection, while CSP \cite{CSP} performs joint channel and space pruning for general CNNs. 
Despite these advances, the assumption remains that channels can be judged independently. But this assumption fails in multimodal detectors, where a channel pruned from one branch may be compensated by the other.

\section{Methodology}
\begin{figure*}[h]
    \centering
    \includegraphics[width=0.95\linewidth]{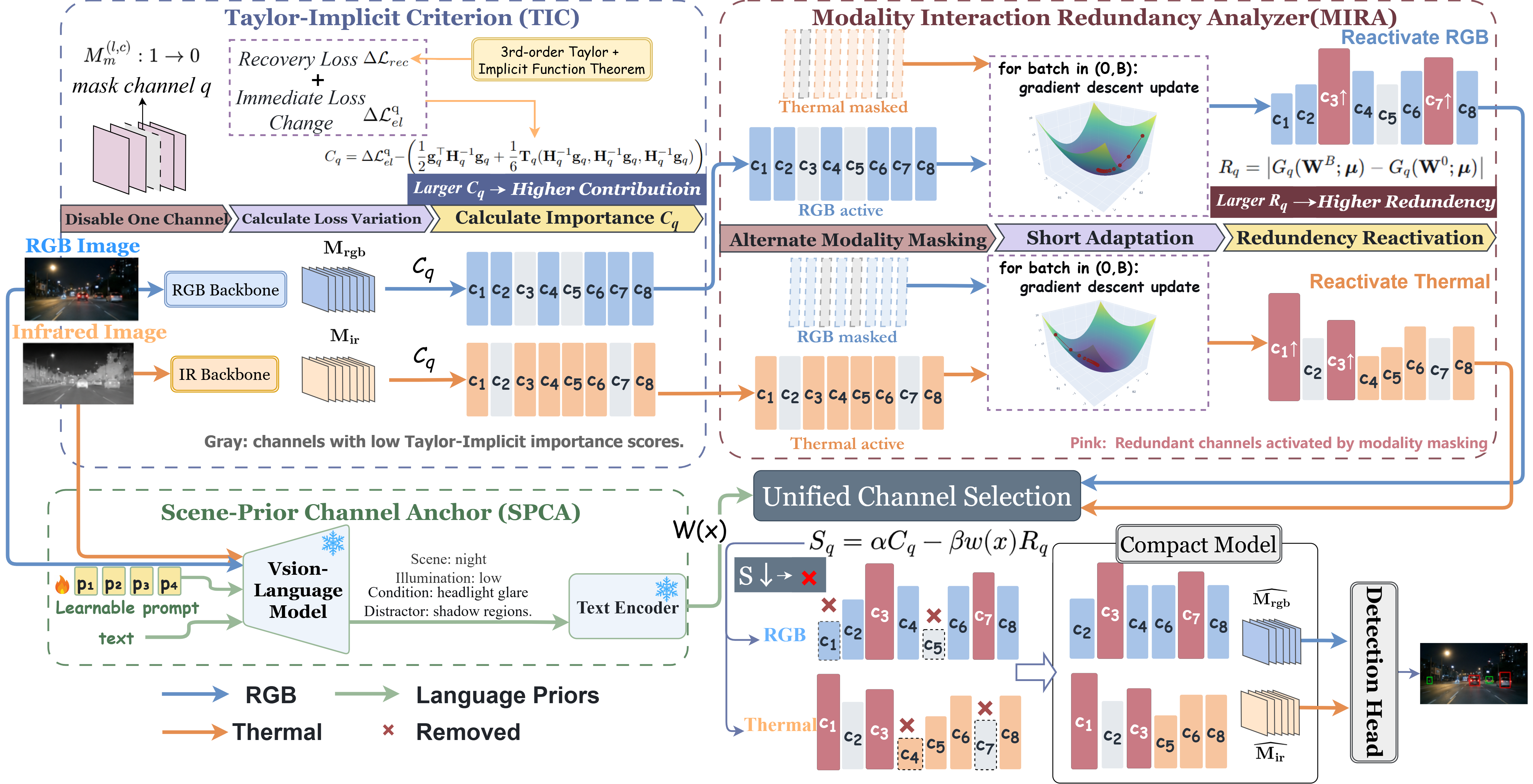}
    \caption{Overveiw of InterPruner: Given multi-scale channels from RGB and infrared backbones, TIC evaluates each channel's contribution $C_q$ via loss variation. MIRA then exposes cross-modal redundancy $R_q$ through alternating masking and compensability assessment. In parallel, SPCA extracts scene semantics from VLM using RGB-IR pairs and prompts to produce a semantic weight 
$w(x)$. These three cues are aggregated into $S_q$
  for channel ranking and pruning, yielding a compact detector.}
    \label{fig:overview}
\end{figure*}
\subsection{Preliminaries}
Let $x_i^{rgb}$ and $x_i^{ir}$ denote aligned RGB and infrared images, and $y_i$ be the detection annotation. 
$\mathbf{W}$ and $\mathbf{M}$ denote the model weights and channel-wise binary masks respectively.
Therefore, a RGB-infrared detector is formulated as:
\begin{equation}
    \hat{y}_i = f(x_i^{rgb}, x_i^{ir}; \mathbf{W}, \mathbf{M}).
\end{equation}
For modality $m \in \{rgb, ir\}$, we define $\mathbf{W}_m = \{\mathbf{W}_m^{(l)}\}_{l=1}^L$ and $\mathbf{M}_m = \{M_m^{(l,c)} \in \{0,1\}\}_{l=1,c=1}^{L,\;C_m^{(l)}}$, with $C_m^{(l)}$ being the number of output channels in layer $l$. 
Each mask $M_m^{(l,c)}=0$ indicates that the $c$-th channel in the $l$-th layer of branch $m$ is pruned during inference.

Given a pretrained detector, we seek an optimal compact mask $\widehat{\mathbf{M}}$ under pruning ratio $\rho$, formulated as an optimization over detection loss $\mathcal{L}$ and training loss $\ell$ \cite{TaylorSecondorder}:
\begin{equation}
\begin{aligned}
    \min_{\widehat{\mathbf{M}} \in \mathcal{S}} \; & \mathcal{L}\bigl(\mathbf{W}^*(\widehat{\mathbf{M}}), \widehat{\mathbf{M}}\bigr), \\
    \text{s.t.} \quad & \mathbf{W}^*(\widehat{\mathbf{M}}) = \arg\min_{\mathbf{W}} \; \ell\bigl(\widehat{\mathbf{M}} \odot \mathbf{W}\bigr) + R(\mathbf{W}),
\end{aligned}
\end{equation}
where $R(\cdot)$ is a regularization term, and $\mathcal{S}$ denotes the feasible set of masks satisfying the pruning ratio $\rho$.

\subsection{Taylor-Implicit Criterion}

The original model parameters $\mathbf{W}$ are trained to convergence $\mathbf{W}^*$ under the mask $\mathbf{M}$.
Removing a single channel gives a new mask $\widetilde{\mathbf{M}}$ and its retrained optimum is $\widetilde{\mathbf{W}}^*$. Here, we define two types of loss changes. The immediate loss change $\Delta \mathcal{L}_{el}$ is the post-pruning loss increase without parameter update:
\begin{equation}
    \Delta \mathcal{L}_{el} = \mathcal{L}(\mathbf{W}^*, \widetilde{\mathbf{M}}) - \mathcal{L}(\mathbf{W}^*, \mathbf{M}).
    \label{eq:immediate_loss}
\end{equation}
The real loss change $\Delta \mathcal{L}_{rl}$ is the loss increase after pruning with full retraining to convergence:
\begin{equation}
    \Delta \mathcal{L}_{rl} = \mathcal{L}(\widetilde{\mathbf{W}}^*, \widetilde{\mathbf{M}}) - \mathcal{L}(\mathbf{W}^*, \mathbf{M}).
    \label{eq:real_loss}
\end{equation}

Directly computing $\Delta \mathcal{L}_{rl}$ for each channel is infeasible. By subtracting  Eq.\ref{eq:immediate_loss} and Eq.\ref{eq:real_loss}, $\Delta \mathcal{L}_{rl}$ can be rewritten as:
\begin{equation}
    \Delta \mathcal{L}_{rl} = \Delta \mathcal{L}_{el} + \mathcal{L}(\widetilde{\mathbf{W}}^*, \widetilde{\mathbf{M}}) - \mathcal{L}(\mathbf{W}^*, \widetilde{\mathbf{M}}).
    \label{eq:loss_decomposition}
\end{equation}

Since $\Delta \mathcal{L}_{el}$ can be directly computed from the weights $\mathbf{W}^*$, our goal is to estimate the recovery loss without fine-tuning, which is defined as 
\begin{equation}
    \Delta \mathcal{L}_{rec}=\mathcal{L}(\widetilde{\mathbf{W}}^*, \widetilde{\mathbf{M}}) - \mathcal{L}(\mathbf{W}^*, \widetilde{\mathbf{M}}).
\end{equation}

Let $\Delta \mathbf{W} = \widetilde{\mathbf{W}}^* - \mathbf{W}^*$ denote the parameter displacement after pruning and fine-tuning.
Since $\widetilde{\mathbf{W}}^*$ is a local optimum under $\widetilde{\mathbf{M}}$, it satisfies:
\begin{equation}
    \nabla_{\mathbf{W}} \mathcal{L}(\widetilde{\mathbf{W}}^*, \widetilde{\mathbf{M}}) = \nabla_{\mathbf{W}} \mathcal{L}(\mathbf{W}^* + \Delta \mathbf{W}, \widetilde{\mathbf{M}})=0
    \label{eq:fisrt_order}
\end{equation}

\paragraph{Taylor Expansion}
We expand the real loss function and its gradient (Eq.\ref{eq:fisrt_order}) around $\mathbf{W}^*$ in Eq.\ref{loss_third_taylor} and Eq.\ref{gradient_taylor}, respectively.
\begin{align}
    \mathcal{L}(\widetilde{\mathbf{W}}^*, \widetilde{\mathbf{M}})
    &= \mathcal{L}(\mathbf{W}^* + \Delta \mathbf{W}, \widetilde{\mathbf{M}}) \label{loss_third_taylor} \\
    &= \mathcal{L}(\mathbf{W}^*, \widetilde{\mathbf{M}}) + \mathbf{g}^T \Delta \mathbf{W} + 
    \frac{1}{2} \Delta \mathbf{W}^T \mathbf{H} \Delta \mathbf{W} \nonumber \\
    &\quad + \frac{1}{6} \mathbf{T}(\Delta \mathbf{W}, \Delta \mathbf{W}, \Delta \mathbf{W}) + O(\|\Delta \mathbf{W}\|^4), \nonumber
\end{align}
\begin{multline}
    \nabla_{\mathbf{W}} \mathcal{L}(\mathbf{W}^* + \Delta \mathbf{W}, \widetilde{\mathbf{M}})
    = 0 \\
    = \mathbf{g} + \mathbf{H} \Delta \mathbf{W} + \frac{1}{2} \mathbf{T}(\Delta \mathbf{W}, \Delta \mathbf{W}) + o(\|\Delta \mathbf{W}\|^3).
    \label{gradient_taylor}
\end{multline}
where we assume $\|\Delta \mathbf{W}\|$ is small, so higher-order terms are negligible. Here $\mathbf{g}$, $\mathbf{H}$, and $\mathbf{T}$ are the first-, second-, and third-order derivatives of $\mathcal{L}$ evaluated at $(\mathbf{W}^*, \widetilde{\mathbf{M}})$.

\paragraph{Implicit Function}
By the implicit function theorem, for sufficiently small $\mathbf{g}$, $\Delta \mathbf{W}$ admits a unique power series expansion in $\mathbf{g}$, with $\Delta \mathbf{W}^{(1)}$ and $\Delta \mathbf{W}^{(2)}$ being the first- and second-order terms:
\begin{equation}
    \Delta \mathbf{W} = \Delta \mathbf{W}^{(1)} + \Delta \mathbf{W}^{(2)} + O(\|\mathbf{g}\|^3).
    \label{eq:w_order}
\end{equation}

Substituting Eq.\ref{eq:w_order} into Eq.\ref{gradient_taylor} and collecting terms by order yields:
\begin{align}
    \Delta \mathbf{W}^{(1)} &= -\mathbf{H}^{-1}\mathbf{g}, \label{delta_w1} \\
    \Delta \mathbf{W}^{(2)} &= -\frac{1}{2}\mathbf{H}^{-1}\mathbf{T}(\Delta \mathbf{W}^{(1)}, \Delta \mathbf{W}^{(1)}). \label{delta_w2}
\end{align}

Thus, the second-order approximation of $\Delta \mathbf{W}$ is
\begin{equation}
    \Delta \mathbf{W} \approx -\mathbf{H}^{-1}\mathbf{g} - \frac{1}{2}\mathbf{H}^{-1}\mathbf{T}(\Delta \mathbf{W}^{(1)}, \Delta \mathbf{W}^{(1)}) + O(\|\mathbf{g}\|^3).
    \label{delta_w}
\end{equation}

\paragraph{Taylor-Implicit Criterion}
Combining Eq.\ref{delta_w} with Eq.\ref{loss_third_taylor} and eliminating higher-order terms via Eq.\ref{delta_w1}--\ref{delta_w2} yields:

\begin{equation}
    \Delta \mathcal{L}_{rec}= -\frac{1}{2}\mathbf{g}^T \mathbf{H}^{-1}\mathbf{g} - \frac{1}{6}\mathbf{T}(\mathbf{H}^{-1}\mathbf{g}, \mathbf{H}^{-1}\mathbf{g}, \mathbf{H}^{-1}\mathbf{g})
\end{equation}

Finally, our Taylor-Implicit Criterion is defined as
\begin{equation}
    C_q
    =
    \Delta\mathcal{L}_{el}^{\mathrm{q}}
    -
    \left(
    \frac{1}{2}
    \mathbf{g}_{q}^{\top}\mathbf{H}_{q}^{-1}\mathbf{g}_{q}
    +
    \frac{1}{6}
    \mathbf{T}_{q}(\mathbf{H}_{q}^{-1}\mathbf{g}_{q}, \mathbf{H}_{q}^{-1}\mathbf{g}_{q}, \mathbf{H}_{q}^{-1}\mathbf{g}_{q})
    \right).
    \label{eq:importance_score}
\end{equation}
This criterion estimates each channel's importance via its contribution to the detection loss. Channels with smaller $C_q$ are less critical and prioritized for pruning.

\subsection{Modality Interaction Redundancy Analyzer}
To analyze the role of each channel under cross-modal interaction, we alternately mask all channels in one modality by setting $\mathbf{M}_{rgb}=\mathbf{0}$ or $\mathbf{M}_{ir}=\mathbf{0}$, and observe the loss variation of the other branch after short adaptation. 

\paragraph{Reactivation via Short Adaptation.}
The discrete mask $M_m^{(l,c)} \in \{0,1\}$ controls channel retention but is non-differentiable for gradient-based optimization. Therefore, we introduce a continuous scalar gate $z_q \in [0,1]$ for each channel $q=(m,l,c)$, which scales the corresponding channel weights as $\mathbf{W}_m^{(l,c)}(z_q) = z_q \cdot \mathbf{W}_m^{(l,c)}$. Here, $z_q=1$ preserves the channel while $z_q=0$ effectively disables it.

Take the infrared branch as the active modality for example. Each reactivation measurement starts from $\mathbf{W}^*$, so the two modalities do not affect each other's results. With the RGB branch disabled, the model undergoes $B$-step gradient adaptation on the infrared branch.
\begin{equation}
    \mathbf{W}^{b+1} = \mathbf{W}^{b} - \eta \nabla_{\mathbf{W}} \mathcal{L}\bigl(\mathbf{W}^{b}), b=0,1,\ldots,B-1,
\end{equation}
where $\eta$ is the short-adaptation learning rate. During this procedure, the masked RGB branch is disabled, forcing the active infrared branch to update its parameters. 
Here, underutilized channels originally masked by the RGB branch contribute substantially more to loss reduction.
We term this \emph{reactivation} and it serves as an indicator of cross-modal redundancy.
Strong reactivation reveals channels that were suppressed, while essential channels show little change. 

\paragraph{Channel Redundancy Measurement.}
We further quantify the loss variation during the $B$-batch adaptation. $G_q$ is defined as the directional derivative of the detection loss with respect to $z_q$. This value reflects each channel's importance under the current mask configuration:
\begin{equation}
G_q(\mathbf{W})
=
\left.
\frac{\partial \mathcal{L}(\mathbf{W}, z_q)}
{\partial z_q}
\right|_{z_q=1}
=
\left\langle
\mathbf{W}_m^{(l,c)},
\frac{\partial \mathcal{L}(\mathbf{W})}
{\partial \mathbf{W}_m^{(l,c)}}
\right\rangle.
\label{eq:channel_gradient}
\end{equation}

Given a set of $B$ batches, we compute the variation of $G_q$ as redundency measurement.
\begin{equation}
    R_q = \frac{1}{B} \left| \sum_{b=1}^B \left( G_q(\mathbf{W}^{B}) - G_q(\mathbf{W}^{0}) \right) \right|.
    \label{eq:reactivation_score}
\end{equation}
A large $R_q$ indicates that the channel becomes more responsive when the opposite modality is absent. This suggests its functionality is suppressed in the original dual-stream model. Conversely, a small $R_q$ implies that the channel's contribution remains largely unchanged regardless of whether the other modality is present. This indicates that it carries modality-specific information and cannot be easily compensated. Therefore, $R_q$ serves as a reliable indicator of cross-modal redundancy. Channels with higher reactivation scores are more dispensable and should be prioritized for pruning.

\subsection{Scene-Prior Channel Anchor}
The relative reliability of RGB and thermal modalities varies with imaging conditions. To adapt pruning decisions accordingly, we introduce a Scene-Prior Channel Anchor (SPCA) that extracts structured scene knowledge from each image pair and converts it into channel-wise pruning weights. SPCA is implemented as a lightweight MLP and applied only to intermediate layers, where effective cross-modal fusion and task-relevant visual pattern recognition primarily occur.

\begin{figure}[t]
    \centering
    \includegraphics[width=1\linewidth]{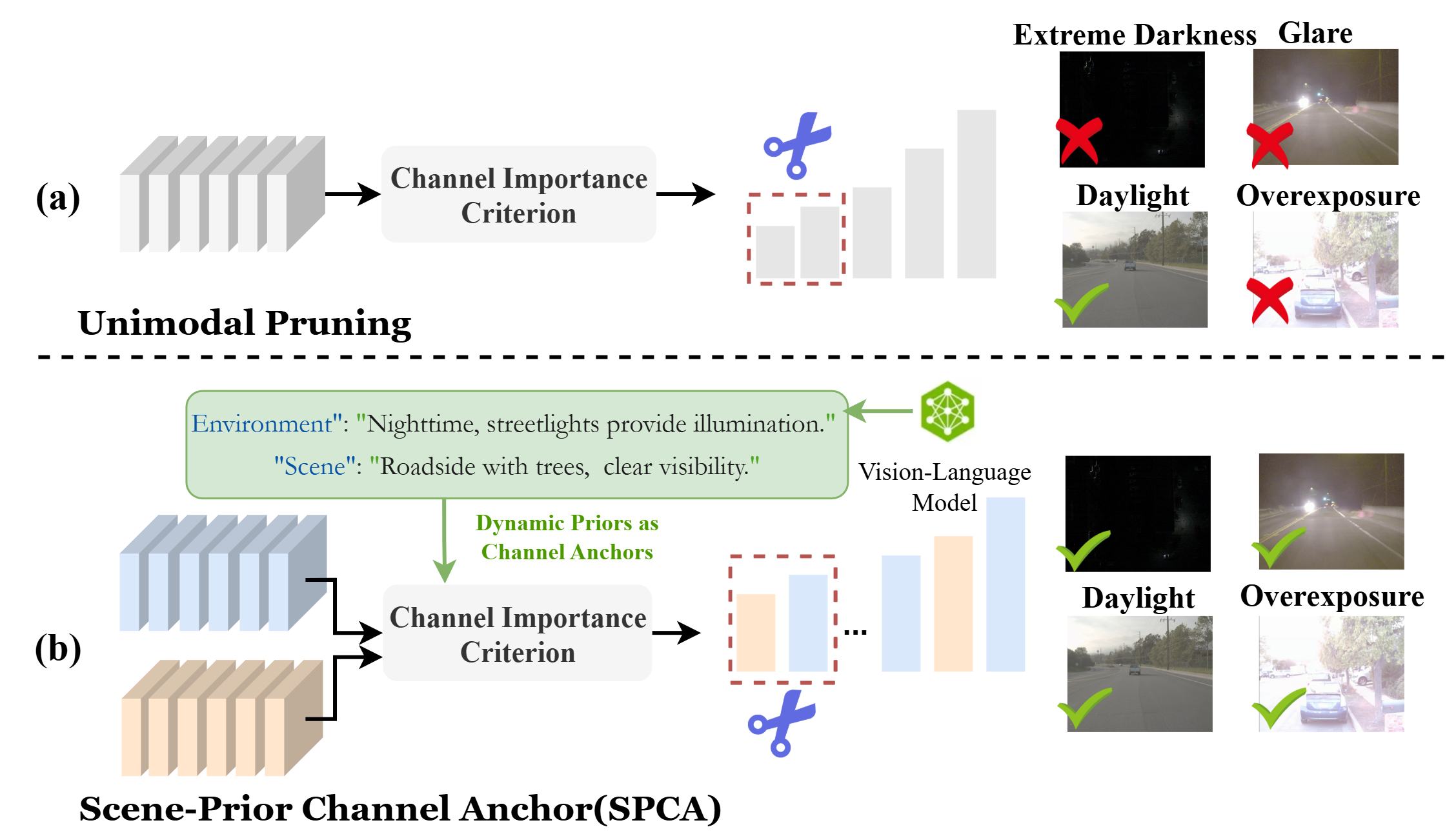}
    \caption{Comparison of the Pruning Method.  While existing methods are suboptimal and fragile under varying lighting conditions, our SPCA enables adaptive pruning decisions conditioned on scene context.}
    \label{fig:placeholder}
\end{figure}

\paragraph{Modality-aware prompt learning.}
For each aligned RGB-Infrared pair, we use a frozen Qwen2.5-VL model with a fixed text prompt to generate structured scene descriptions. Additionally, $K$ learnable soft prompts $\mathbf{P}\in\mathbb{R}^{K\times d_{\mathrm{Q}}}$ is also inserted into the multimodal input to make the encoding sensitive to modality reliability.

We optimize $\mathbf{P}$ using modality-difference supervision from a frozen CLIP image encoder. Specifically, we define the RGB-Infrared discrepancy direction $\mathbf{v}_i^-$ and common direction $\mathbf{v}_i^+$ from the normalized CLIP embeddings of each pair. 
Let $\mathbf{H}_i^P$ be the prompt tokens' hidden states. Their pooled representation is projected into the CLIP space via a linear layer $A$ to obtain $\mathbf{u}_i$. 
The prompt-learning objective is:
\begin{equation}
    \mathcal{L}_{\mathrm{prompt}} = \frac{1}{B}\sum_{i=1}^{B} (1-\mathbf{u}_i^\top \mathbf{v}_i^-) + \lambda_{1} \frac{1}{B}\sum_{i=1}^{B} (\mathbf{u}_i^\top \mathbf{v}_i^+)^2 + \lambda_2\mathcal{L}_{\mathrm{con}},
\end{equation}
which aligns prompts with RGB-thermal discrepancy, suppresses common information, and separates different pairs, with only $\mathbf{P}$ and $A$ trainable.

\paragraph{Structured semantic encoding.}
With the learned prompt $\mathbf{P}^{*}$, Qwen2.5-VL generates a structured description for each RGB-IR pair:
\begin{equation}
    \mathcal{D}_{i} = G_{\mathrm{Qwen}}(x_i^{r}, x_i^{t}; \mathbf{P}^{*}) = \{d_i^{\mathrm{env}}, d_i^{\mathrm{scene}}, d_i^{\mathrm{obj}}, d_i^{\mathrm{therm}}\},
\end{equation}
covering illumination, scene context, object-level evidence, and thermal properties. These four text fields are encoded by the CLIP text encoder and averaged along the field dimension to obtain a scene-level semantic embedding $\mathbf{z}_i \in \mathbb{R}^{768}$.

\paragraph{Channel-wise pruning modulation.}

Prompt learning and semantic description generation are performed offline before pruning calibration. The core of SPCA is to transform scene representation into channel-wise weights that modulate the pruning criterion. Let $b=(m,l)$ denote layer $l$ in modality branch $m\in\{r,t\}$, and let $C_b$ be its number of output channels. We implement the modulation network as a lightweight two-layer MLP that maps the 768-dimensional CLIP text embedding $\mathbf{z}_i$ to channel-wise weights:

\begin{equation}
    \boldsymbol{\delta}_{i,b}
    =
    \tanh
    \left(
    \mathbf{W}_2
    \operatorname{ReLU}
    \left(
    \mathbf{W}_1 \mathbf{z}_i
    +
    \mathbf{b}_1
    \right)
    +
    \mathbf{b}_2
    \right)
    \in\mathbb{R}^{C_b}.
\label{eq:channel_residual}
\end{equation}

To prevent the semantic module from uniformly scaling all channels in a branch, we center and normalize the residual:
\begin{equation}
    \widehat{\boldsymbol{\delta}}_{i,b}
    =
    \frac{
    \boldsymbol{\delta}_{i,b}
    -
    \operatorname{Mean}
    (\boldsymbol{\delta}_{i,b})
    }{
    \max\left(
    1,
    \left\|
    \boldsymbol{\delta}_{i,b}
    -
    \operatorname{Mean}
    (\boldsymbol{\delta}_{i,b})
    \right\|_{\infty}
    \right)
    }.
\label{eq:centered_residual}
\end{equation}

The channel-wise semantic weights are defined as follow, where $\gamma$ controls the modulation range, yielding $\mathbf{w}_{i,b}\in[1-\gamma,1+\gamma]^{C_b}$.
\begin{equation} 
    \mathbf{w}_{i,b}
    =
    \mathbf{1}
    +
    \gamma
    \widehat{\boldsymbol{\delta}}_{i,b},
\label{eq:semantic_weight}
\end{equation}

The lightweight MLP is optimized via the detection objective with differentiable channel masks. Since physical pruning requires a fixed architecture, we average the sample-dependent weights over the calibration set:
\begin{equation}
    \overline{w}_{m,l,c}
    =
    \frac{1}{N_{\mathrm{cal}}}
    \sum_{i=1}^{N_{\mathrm{cal}}}
    w_{i,m,l,c}.
\label{eq:average_semantic_weight}
\end{equation}
The averaged weights are substituted into Eq.~\eqref{eq:semantic_pruning_score} to determine the final channel ranking.

\subsection{Unified Channel Selection}
\begin{table*}[t]
    \centering
    \setlength{\tabcolsep}{18pt}
    \renewcommand{\arraystretch}{0.95}  
    \begin{tabular}{l ccc ccc}
        \toprule
        \multirow{2}{*}{\textbf{Channel Criterion}}
        & \multicolumn{3}{c}{\textbf{DroneVehicle}}
        & \multicolumn{3}{c}{\textbf{FLIR}} \\
        \cmidrule(lr){2-4}
        \cmidrule(lr){5-7}
        & 30\% & 50\% & 70\%
        & 30\% & 50\% & 70\% \\
        \midrule
        $\ell_1$ Norm Pruning~\cite{L12016pruning}
        & 67.6 & 66.3 & 46.7
        & 64.4 & 56.4 & 41.5 \\

        Network Slimming~\cite{NetworkSliming}
        & 67.8 & 66.0 & 64.8
        & 49.8 & 49.2 & 44.2 \\

        Intra-Fusion~\cite{Intra-Fusion:theus2024metapruningoptimaltransport}
        & 75.8 & 64.6 & 47.9
        & 66.6 & 48.3 & 31.3 \\

        BilevelPruning~\cite{BilevelPruning}
        & 77.2 & 72.3 & 70.4
        & 75.6 & 74.8 & 73.5 \\

        REPrune~\cite{RepPrune}
        & 67.0 & 63.5 & 59.4
        & 55.3 & 53.7 & 53.1 \\

        IRPruneDeXt~\cite{IRPruneDeXt}
        & 76.7 & 76.1 & 63.7
        & 73.1 & 72.5 & 45.4 \\

        Probabilistic Pruning~\cite{Probabilistic}
        & 74.2 & 71.8 & 65.3
        & 70.5 & 67.1 & 58.7 \\

        CSP~\cite{CSP}
        & 78.5 & 76.2 & 72.1
        & 76.3 & 73.9 & 68.4 \\

        HOT-MKS~\cite{2026lightweight}
        & 80.4 & 78.0 & 73.4
        & 74.1 & 70.3 & 63.3 \\

        \midrule
        \textbf{InterPruner (Ours)}
        & \textbf{81.0} & \textbf{80.3} & \textbf{79.0}
        & \textbf{77.5} & \textbf{76.7} & \textbf{72.6} \\
        \bottomrule
    \end{tabular}
    \caption{Comparison of mAP$_{50}$ (\%) across different channel importance criteria and pruning ratios on the DroneVehicle and FLIR datasets. The best results are highlighted in bold.}
    \label{tab:sota}
\end{table*}
With the channel contribution $C_q$ from TIC and the cross-modal redundancy $R_q$ from MIRA established, we now integrate them into a unified importance score. A channel should be preserved if it is highly contributive to the detection task, and penalized if it is redundantly substitutable by the other modality. Besides, we introduce a scene-conditioned weight $w_{i,m,l,c}$, predicted by our SPCA module, which adaptively adjusts the redundancy penalty according to the semantic context. The final importance score for channel $c$ in layer $l$ of modality branch $m$ is formulated as:

\begin{equation}
    S_{i,m,l,c}
    =
    \alpha \, \overline{C}_{m,l,c}
    -
    \beta \, w_{i,m,l,c} \, \overline{R}_{m,l,c},
\label{eq:semantic_pruning_score}
\end{equation}
where $\overline{C}_{m,l,c}$ and $\overline{R}_{m,l,c}$ are the normalized contribution and redundancy scores, and $\alpha,\beta$ are fixed global weights. The sample-dependent $w_{i,m,l,c} \in [1-\gamma, 1+\gamma]$ modulates the redundancy penalty: $w_{i,m,l,c} > 1$ amplifies the penalty, making the channel more likely to be pruned, while $w_{i,m,l,c} < 1$ reduces it, allowing retention even if redundant. 

With the unified importance score $S_{i,m,l,c}$, we jointly prune the RGB and infrared backbones. For each modality, channels in each layer are sorted in ascending order of their scores, with lower scores indicating higher dispensability. Under an equal-width constraint, we prune the same number of channels from both modalities within each layer pair to preserve structural compatibility for subsequent fusion modules. To maintain robustness, we pre-select the top 10\% of channels with the highest contribution scores \(C_q\) and exclude them from pruning. The global pruning budget is then allocated across all layers via a greedy strategy. At each step, the lowest-scored available channel across all layers is pruned, and the next channel from the same layer enters the candidate pool, until the target ratio \(\rho\) is met. This enables global competition across layers while maintaining per-layer RGB/IR balance. Thus, we obtain the compact model.

\section{Experiments}
\subsection{Datasets}
\paragraph{FLIR.} FLIR \cite{Flir} focuses on complex outdoor driving scenarios, comprising 10,228 images (8,862 training, 1,366 testing) with annotations for Person, Car, Bicycle. It is characterized by crowded streets, significant scale variations, and cluttered backgrounds. These conditions pose a substantial challenge to the model’s ability to preserve structural details and distinguish objects in dense environments.

\paragraph{DroneVehicle.}  DroneVehicle\cite{DroneVehicle} consists of 56,878 image pairs collected by UAVs, featuring an aerial perspective. It covers five vehicle categories (Car, Truck, Bus, Van, Freight-Car) and provides oriented bounding box annotations. The dataset introduces unique challenges such as small object scales, high density, and complex background textures, setting a high standard for evaluating the adaptability of multimodal detectors in aerial surveillance scenarios.

\subsection{Implementation Details}
We evaluate all models using mAP and mAP$_{50}$, and report their FLOPs and parameter counts. We implement InterPruner using the Ultralytics YOLOv8 framework with a pretrained C2DFF model~\cite{C2DFF-Net} on NVIDIA RTX 3090 GPUs. For TIC and MIRA, we perform 8-batch adaptation with a learning rate of \(1 \times 10^{-3}\). For SPCA, the learnable 16-token Qwen2.5-VL prompt is optimized using AdamW at \(1 \times 10^{-2}\), with a frozen CLIP ViT-L/14 encoder providing modality-difference supervision. The pruned detector is fine-tuned with SGD at \(1 \times 10^{-3}\). The balancing coefficients \(\alpha\) and \(\beta\) are both set to 1. For baselines requiring fine-tuning, we use \(1 \times 10^{-4}\) for 80 epochs at 30\% and 50\% pruning, and \(1 \times 10^{-3}\) for 70\% pruning.

\subsection{Main Results}

We conduct extensive experiments to validate the effectiveness of our proposed InterPruner. As seldom pruning method is specifically designed for the infrared-visible dual-modality setting, we adapt existing single-modality pruning methods to serve as baselines. This allows us to perform a comprehensive analysis and ultimately evaluate the superiority of our cross-modal pruning strategy. Specifically, we compare against L1 Norm, Network Slimming\cite{NetworkSliming}, Intra-Fusion\cite{Intra-Fusion:theus2024metapruningoptimaltransport}, BilevelPruning\cite{BilevelPruning}, REPrune\cite{RepPrune}, IRPruneDecX\cite{Intra-Fusion:theus2024metapruningoptimaltransport}, and CSP \cite{CSP}.

\paragraph{Comparison with Single Model Pruning Criteria.}
As shown in Table~\ref{tab:sota}, directly applying existing single-branch pruning criteria to RGB-infrared detectors leads to performance degradation, especially under large pruning ratios. 
Magnitude-based methods (\(\ell_1\) Norm Pruning~\cite{L12016pruning}, Network Slimming~\cite{NetworkSliming}) perform the worst in multimodal scenarios. This confirms that single-channel weights or scaling factors cannot discern cross-modal complementarity, thus inevitably remove vital branches.

Recent works have explored structured pruning from various perspectives. Similarity-based approaches (BilevelPruning~\cite{BilevelPruning}, REPrune~\cite{RepPrune}) achieve 72.3\% on DroneVehicle dataset at 50\% pruning, demonstrating moderate performance. However, they merely focus on intra-modal geometric redundancy while completely ignoring inter-modal dependencies. Gradient-based method~\cite{2026lightweight} outperforms other baselines (80.4\% at 30\% pruning on DroneVehicle), proving that Taylor expansion effectively retains loss-sensitive channels. Yet, localized gradient estimation proves brittle under extreme pruning, as evidenced by a 6.8\% mAP\(_{50}\) drop on FLIR dataset at the 70\% ratio. Reconstruction-based methods (Intra-Fusion~\cite{Intra-Fusion:theus2024metapruningoptimaltransport}) partially compensate for pruning-induced loss through feature reconstruction, but this comes at the cost of substantial computational overhead and limited scalability to high pruning ratios. Although some of these single-modality metrics perform well at low pruning ratios, they fundamentally lack the capacity to model cross-modal complementarity and thus fail under aggressive pruning. Therefore, InterPruner jointly assesses channel importance across modalities, achieving state-of-the-art performance across all settings and maintaining robustness even at 70\% pruning.
\begin{figure}
    \centering
    \includegraphics[width=1\linewidth]{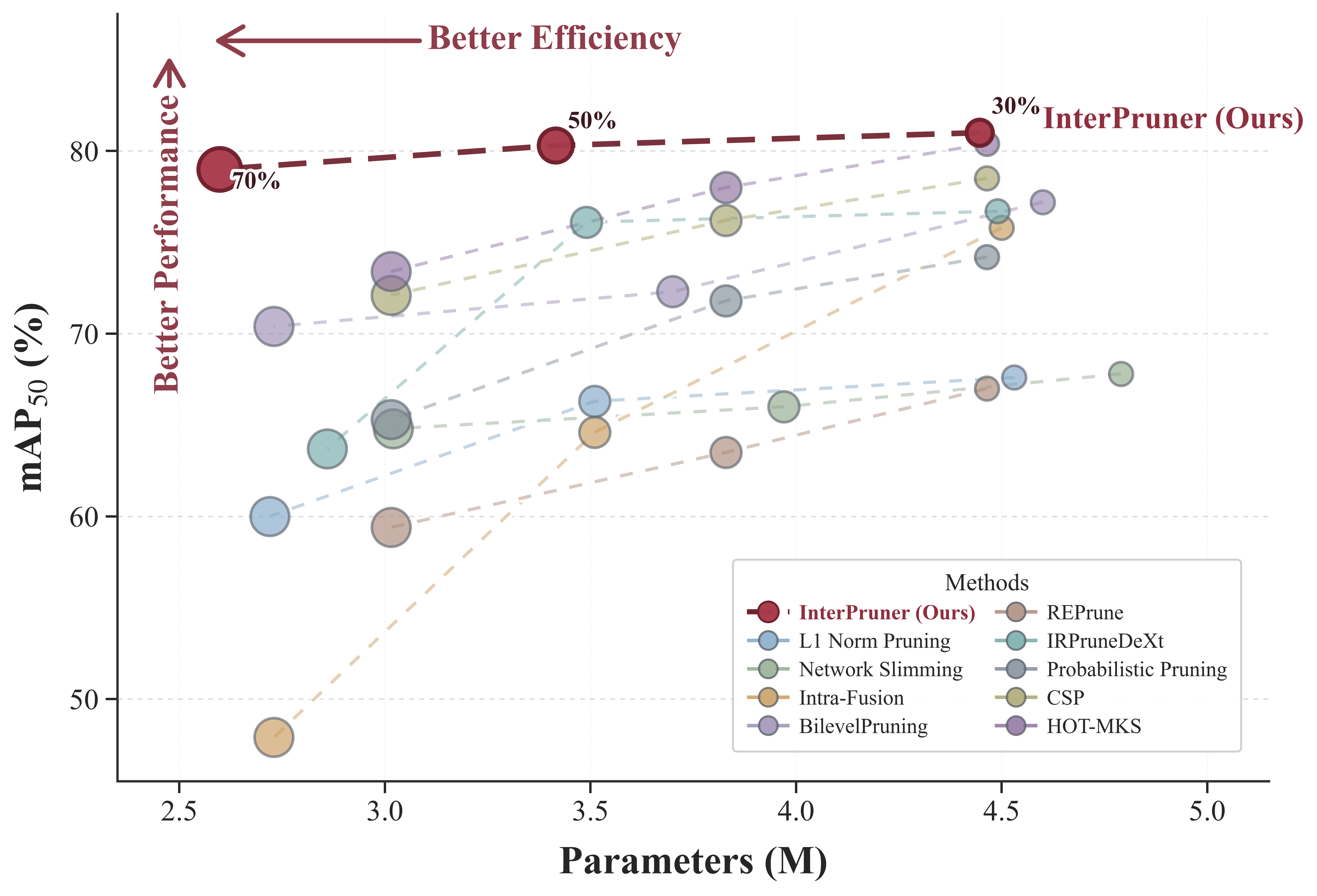}
    \caption{Comparison with other pruning methods on the DroneVehicle dataset in terms of mAP\(_{50}\) and model size.}
    \label{fig:placeholder}
\end{figure}
\begin{figure}
    \centering
    \includegraphics[width=0.8\linewidth]{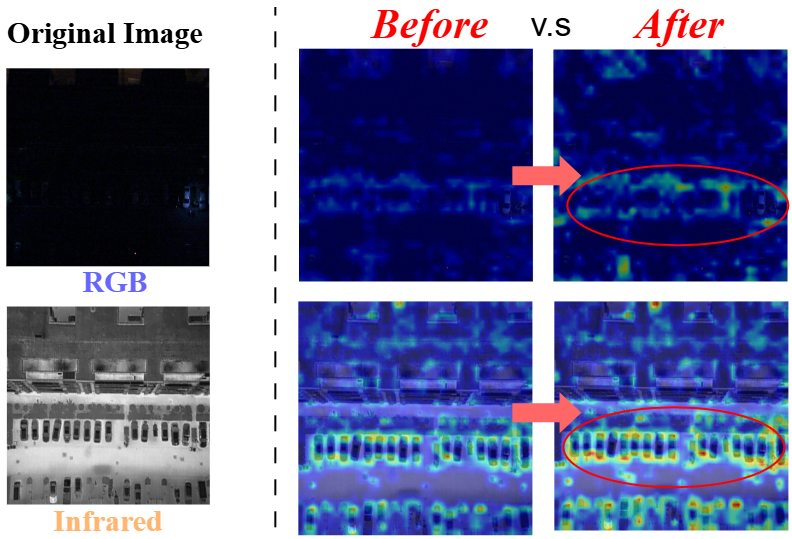}
    \caption{Feature visualizations of the RGB and Infrared branches before and after pruning under extremely dark condition. The pruned model focuses on more salient regions in both modalities, showing the effectiveness of InterPruner.}
    \label{fig:placeholder}
\end{figure}

\paragraph{Comparison with the Original Model under Varying Pruning Ratios}
Table~\ref{tab:robust} presents the performance and complexity of InterPruner across varying pruning ratios on FLIR and DroneVehicle. As pruning intensifies, parameter and FLOP counts decline substantially, yet mAP\(_{50}\) exhibits only negligible degradation on both datasets. This underscores InterPruner's insensitivity to specific pruning levels, achieving a flexible accuracy-efficiency trade-off and affirming the robustness of the proposed strategy.
\begin{table}[t]
    \centering
    \setlength{\tabcolsep}{4.2pt}
    \renewcommand{\arraystretch}{1}
    \resizebox{\columnwidth}{!}{
    \begin{tabular}{c c c c c c}
        \toprule
        Dataset 
        & Pruning Ratio 
        & mAP$_{50}$ (\%)
        & mAP(\%)
        & Params (M) 
        & FLOPs (G) \\
        \midrule

        \multirow{4}{*}{FLIR}
        & 30\%
        & 77.5
        & 41.5
        & 4.695
        & 11.4 \\

        & 50\%
        & 76.7
        & 41.4
        & 3.696
        & 10.4 \\

        & 70\%
        & 72.6
        & 38.7
        & 2.767
        & 7.9 \\

        & -
        & 76.8
        & 40.8
        & 6.580
        & 14.6 \\
        \midrule

        \multirow{4}{*}{DroneVehicle}
        & 30\%
        & 81.0
        & 60.4
        & 4.446
        & 9.5 \\

        & 50\%
        & 80.3
        & 59.7
        & 3.415
        & 8.1 \\

        & 70\%
        & 79.0 
        & 58.4
        & 2.598 
        & 7.2 \\

        & -
        & 80.2
        & 59.8
        & 6.580
        & 14.6 \\
        \bottomrule
    \end{tabular}
    }
    \caption{Performance comparison of pruned models at different ratios and the original model.}
    \label{tab:robust}
\end{table}

\begin{table}[t]
\centering
\small
\setlength{\tabcolsep}{2.5pt}  
\renewcommand{\arraystretch}{1.1}
\begin{tabular}{c c c c c c c c c}  
\hline
\multirow{2}{*}{TIC} & \multirow{2}{*}{MIRA} & \multirow{2}{*}{SPCA}
& \multicolumn{2}{c}{30\%} & \multicolumn{2}{c}{50\%} & \multicolumn{2}{c}{70\%} \\
\cline{4-5} \cline{6-7} \cline{8-9}
 & & & mAP$_{50}$ & mAP & mAP$_{50}$ & mAP & mAP$_{50}$ & mAP \\
\hline
\checkmark & & & 72.4 & 39.2 & 70.2 & 38.1 & 62.7 & 33.7 \\
\checkmark & \checkmark & & 76.3 & 41.2 & 74.4 & 40.0 & 67.1 & 35.4 \\
\checkmark & \checkmark & \checkmark & \textbf{77.5} & \textbf{41.5} & \textbf{76.7} & \textbf{41.4} & \textbf{72.6} & \textbf{38.7} \\
\hline
\end{tabular}
\caption{Ablation study of the proposed components on FLIR under different pruning ratios. Results are percentage.}
\label{tab:ablation_components}
\end{table}
\subsubsection{Component Ablations.}
The ablation results are summarized in Table~\ref{tab:ablation_components}. TIC effectively estimates channel importance using high-order expansion and establishes a moderate baseline across pruning ratios. MIRA further boosts the performance and the gain grows with sparsity. We attribute this to its explicit cross-modal modeling, which prevents the removal of channels that are individually less important but provide complementary information to the other modality. Adding SPCA improves mAP\(_{50}\) by 5.5\% at 70\% pruning, and the improvement widens as pruning becomes aggressive. This suggests SPCA adapts selection to scene-dependent distributions, which is critical at high sparsity, where fixed criteria often fail. Ultimately, the full model preserves high precision across pruning ratios, confirming that TIC, MIRA, and SPCA collectively address dynamic channel importance.
\begin{table}[t]
    \centering
    \setlength{\tabcolsep}{8pt}
    \renewcommand{\arraystretch}{1.1}
    \begin{tabular}{l c c c}
        \toprule
        TIC Approximation & 30\% & 50\% & 70\% \\
        \midrule
        First-order Taylor  & 38.2 & 37.9 & 35.1 \\
        Second-order Taylor & 39.4 & 39.1 & 36.4 \\
        Third-order Taylor  & \textbf{41.5} & \textbf{41.4} & \textbf{38.7} \\
        \midrule
        Params (M) & 4.695 & 3.696 & 2.767 \\
        \bottomrule
    \end{tabular}
    \caption{Effect of Taylor expansion order in TIC on FLIR under different pruning ratios. Results are reported in mAP(\%).}
    \label{tab:taylor_order}
\end{table}
\subsubsection{Analysis of Taylor-Implicit Criterion.}
We investigate the influence of different Taylor approximation orders in TIC in Table~\ref{tab:taylor_order}. First-order Taylor only captures the immediate gradient response after channel removal, and it exhibits the poorest performance across all pruning ratios, yielding only 38.2\% mAP at 30\% channel pruning. While introducing the second-order term effectively improves the results by considering local curvature, it remains insufficient to model the nonlinearity inherent in our compensation-aware strategy. By incorporating the third-order term, TIC better models the nonlinear loss variation caused by channel removal and parameter compensation, leading to the best performance under all pruning ratios. These results confirm that high-order Taylor expansion outperforms lower-order alternatives by effectively modeling the loss induced by pruning and compensation, ensuring reliable channel importance estimation.

\subsubsection{Sensitivity Analysis of Hyper-parameters in Unified Channel Selection}
\begin{figure}[!htb]
    \centering
    \includegraphics[width=1\linewidth]{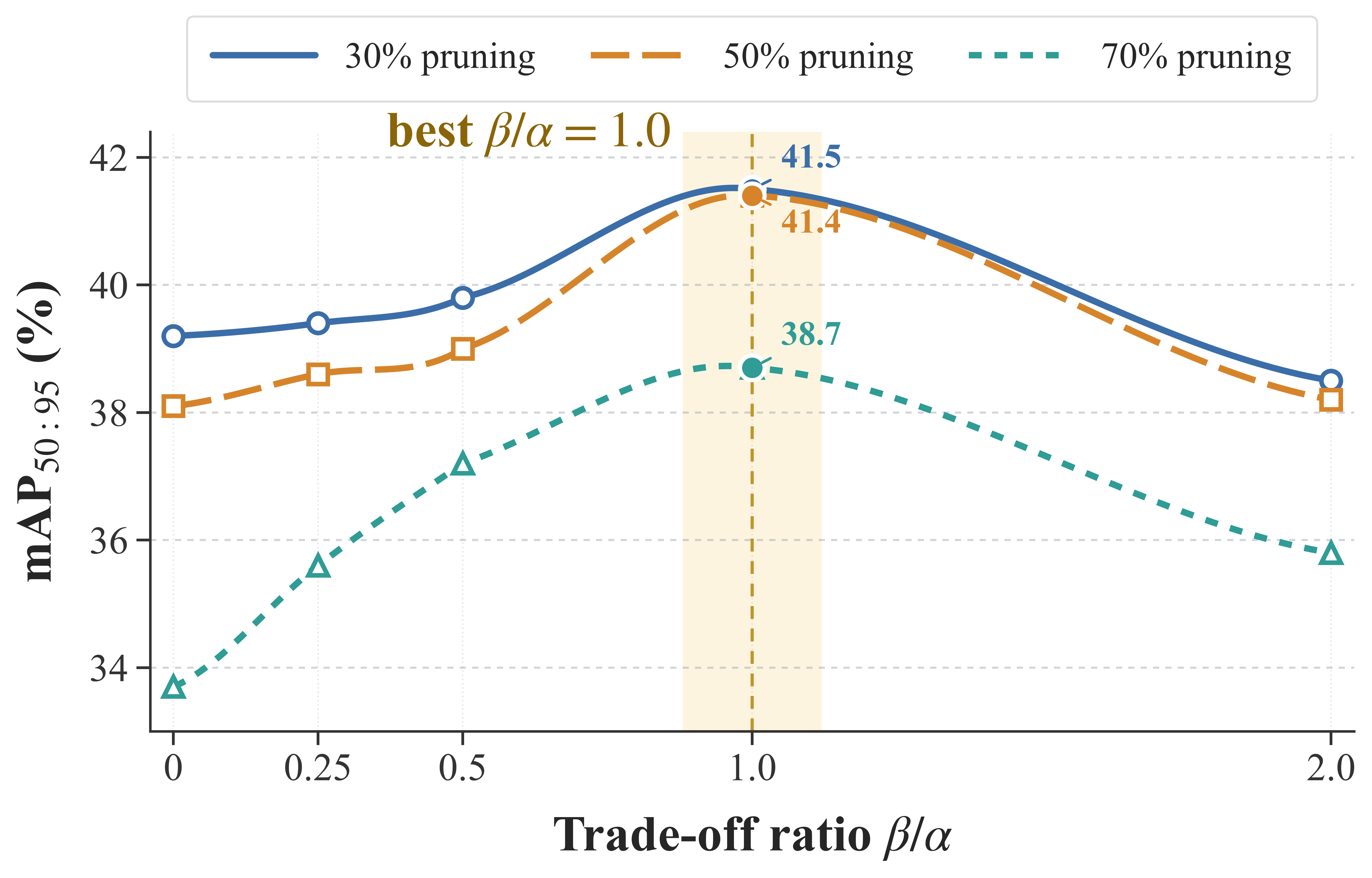}
    \caption{Visualization of the $\beta/\alpha$ ablation study in Table~\ref{tab:hyper_alpha_beta}.}
    \label{fig:curve}
\end{figure}

We ablate the trade-off ratio \(\beta/\alpha\) in Eq.~\eqref{eq:semantic_pruning_score} on the FLIR dataset under three distinct pruning ratios.
As shown in Fig.~\ref{fig:curve}, the mAP consistently peaks at \(\beta/\alpha=1.0\), reaching 41.5\%, 41.4\%, and 38.7\% respectively. This unified peak shows the optimal ratio is pruning-agnostic. In fact, deviations from this setting cause performance degradation. A small ratio ignores compensation potential, while a large one overemphasizes it and may discard vital features, especially under limited capacity. Therefore, setting \(\beta/\alpha=1.0\) ensures an optimal trade-off between information retention and pruning efficiency, validating that our criterion remains effective even under aggressive pruning.

\begin{table}[t]
\centering
\small
\setlength{\tabcolsep}{6pt}
\renewcommand{\arraystretch}{1}
\begin{tabular}{l c c c c c c}
\toprule
\multirow{2}{*}{\(\beta/\alpha\)} & \multicolumn{2}{c}{30\%} & \multicolumn{2}{c}{50\% } & \multicolumn{2}{c}{70\%} \\
\cmidrule(lr){2-3} \cmidrule(lr){4-5} \cmidrule(lr){6-7}
 & mAP\(_{50}\) & mAP & mAP\(_{50}\) & mAP & mAP\(_{50}\) & mAP \\
\midrule
0.0  & 72.4 & 39.2 & 70.2 & 38.1 & 62.7 & 33.7 \\
0.25 & 73.5 & 39.4 & 72.2 & 38.6 & 67.1 & 35.6 \\
0.50 & 74.3 & 39.8 & 73.6 & 39.0 & 69.7 & 37.2 \\
1.00 & \textbf{77.5} & \textbf{41.5} & \textbf{76.7} & \textbf{41.4} & \textbf{72.6} & \textbf{38.7} \\
2.00 & 72.0 & 38.5 & 71.2 & 38.2 & 67.5 & 35.8 \\
\bottomrule
\end{tabular}
\caption{Ablation study of the \(\beta/\alpha\) on the FLIR dataset across different pruning ratios. Results are percentage.}
\label{tab:hyper_alpha_beta}
\end{table}

\section{Conclusion}
In this paper, we propose InterPruner, an interactive structured pruning framework for RGB-infrared object detectors. The framework first assesses channel importance via a Taylor-Implicit Criterion, then identifies cross-modal redundancy through a Modality Interaction Redundancy Analyzer, and finally leverages a Scene-Prior Channel Anchor that adapts pruning to semantic context.
This pruning paradigm offers new insights for lightweight multimodal perception, with natural extensions to other resource-constrained applications, such as heterogeneous segmentation, medical imaging, and autonomous systems, where preserving cross-modal information with minimal overhead is critical.


\newpage
{\small
\bibliographystyle{ieee_fullname}
\bibliography{egbib}
}

\end{document}